# The Two Word Test: A Semantic Benchmark for Large Language Models

Nicholas Riccardi and Rutvik H. Desai

University of South Carolina

Department of Psychology

## Abstract

Large Language Models (LLMs) have shown remarkable abilities recently, including passing advanced professional exams and demanding benchmark tests. This performance has led many to suggest that they are close to achieving humanlike or "true" understanding of language, and even Artificial General Intelligence (AGI). Here, we provide a new open-source benchmark that can assess semantic abilities of LLMs using two-word phrases using a task that can be performed relatively easily by humans without advanced training. Combining multiple words into a single concept is a fundamental aspect of human language and intelligence. The test requires meaningfulness judgments of 1768 noun-noun combinations that have been rated as meaningful (e.g., *baby boy*) or not meaningful (e.g., *goat sky*) by 150 human participants. We provide versions of the task that probe meaningfulness ratings on a 0-4 scale as well as binary judgments. We conducted a series of experiments using the TWT on GPT-4, GPT-3.5, and Bard, with both versions. Results demonstrate that, compared to humans, all models perform poorly at rating meaningfulness of these phrases. GPT-3.5 and Bard are also unable to make binary discriminations between sensible and nonsense phrases, with both models consistently judging highly nonsensical phrases as making sense. GPT-4 makes a substantial improvement in binary discrimination of combinatorial phrases but is still significantly worse than human performance. The TWT can be used to understand the limitations and weaknesses of current LLMs, and potentially improve them. The test also reminds us that caution is warranted in attributing "true understanding" or AGI to LLMs. TWT is available at: https://github.com/NickRiccardi/two-word-test

## Introduction

Large Language Models (LLMs; also called Large Pre-Trained Models or Foundation Models) (Bommasani et al. 2021) are deep neural networks with billions or trillions or parameters that are trained on massive natural language corpora. They have shown remarkable and surprising abilities spanning many different tasks. Some examples include the ability to pass examinations required for advanced degrees, such as those in law (Choi et al. 2023), business (Terwiesch 2023), and medicine (Kung et al. 2023). Strong performance on benchmarks such as General Language Understanding Evaluation (GLUE) and its successor (SuperGLUE) have also been obtained (Brown et al. 2020, Chowdhery et al. 2022). Bubeck et al. (2023) investigated an early version of GPT-4, and reported that it can solve difficult tasks in mathematics, coding, vision, medicine, law, and psychology, music, and exhibited "mastery of language." With such breadth of human-level (or better) performance, they suggested that it shows "sparks" of Artificial General Intelligence (AGI).

Such achievements have led many researchers to conclude that LLMs have achieved or are close to achieving real or humanlike understanding of language. Others remain skeptical. A recent survey (Michael et al. 2022) asked active researchers whether such models, trained only on text, could in principle understand natural language someday. About half (51%) agreed, while other half (49%) disagreed. This stark divide is closely tied to the question of what constitutes true understanding and is subject of intense debate (Michell and Karkauer 2023).

The skeptics have pointed out examples where LLMs produce less-than-satisfactory performance. Hallucinations (Lee et al. 2018, Raunak et al. 2021), inaccurate number comparisons, and reasoning errors are commonly cited problems, and failures in individual cases are frequently reported (e.g., https://github.com/giuven95/chatgpt-failures). It is argued that while LLMs exhibit *formal* linguistic competence, they lack *functional* linguistic competence, which is the ability to robustly understand and use language in the real world (Mahowald et al. 2023). However, this claim still runs into the problem of how robust understanding is to be measured beyond subjective assessments of the quality of answers in response to prompts. Objective benchmarks are essential here, but as successes and failures of LLMs show, benchmarks that are suitable for measuring human understanding might not be appropriate for assessing LLMs (Choudhury et al. 2022; Gardner et al. 2021; Linzen 2020).

There are philosophical arguments as to why LLMs do not have true or humanlike understanding. For example, LLMs learn words-to-words mappings, but not words-to-world mappings, and hence cannot understand the objects or events that words refer to (Browning and LeCun, 2022). Such arguments aside, formal tests are critical, as that's where "rubber meets the road." If a system can match or surpass human performance in any task thrown at it, the argument that it does not possess *real* understanding, for whatever reason, rings hollow. If an LLM indeed lacks humanlike understanding, one ought to be able to design tests where it performs worse than humans. With such tests, the nebulous definition of "understanding" becomes less of a problem.

Here, we propose and evaluate one such novel benchmark, the Two Word Test (TWT). The test is based on a basic human psycholinguistic ability to understand combinations of two words. The test uses noun-noun combinations (e.g., *beach ball*) and requires discrimination between meaningful and nonsense (e.g., ball beach) combinations. Compared to other types of linguistic compositions, , such as adjective-noun (*big ball*) or verb-noun (*throw ball*),.noun-noun combinations do not offer grammatical assistance in determining meaningfulness. One can determine that *ball red* is not a meaningful phrase, because noun-adjective is not a valid word order in English. The same strategy cannot be used to determine that *ball beach* has low meaningfulness. Some phrases are learned as single units that combine unrelated words (*sea lion*), while others are 'built from the ground up'. *Baby boy* makes sense, and many other words could follow *baby* and the phrase would still be sensible (*clothes*, *girl*, *sister*, etc.). Simply reversing word order of some of these (*clothes baby*) can result in a low-meaningfulness phrase. These unique memory-dependent, semantic, and compositional elements make this test a valuable semantic benchmark for LLMs. A previous study (Graves et al 2013) obtained meaningfulness ratings on these phrases from 150 human participants, which we use here. We report results from three current LLMs (OpenAI's GPT-4 and GPT-3.5-turbo and Google's Bard). Our main contributions are as follows:

- Two Word Test (TWT), a novel open-source benchmark for measuring LLM comprehension using simple two-word phrases. Unlike existing benchmarks, the test does not rely on the ability to do logical reasoning, planning, infer implied content, disambiguate ambiguous words, or solve puzzles or other problems, but relies on combinatorial semantics.

- The TWT measures the ability of LLMs to judge meaningfulness using a Likert scale. We provide a second version, binary TWT (bTWT), which measures binary 'makes sense' or 'nonsense' judgments for each phrase, which is expected to be easier for LLMs.

- A comprehensive statistical comparison using Signal Detection Theory (SDT) metrics and permutation testing, of the performance of GPT-4, 3.5-turbo, and Bard to human data.

- Identification of limitations of current LLMs in language comprehension ability, as a weakness distinct from those in tasks that rely on executive control, such as logical reasoning or puzzle solving.

**Materials and Methods**

**Two Word Test Phrase Generation and Human Rating Collection**

The TWT consists of noun-noun combinations and human meaningfulness ratings collected as part of behavioral and neuroimaging experiments conducted by Graves and colleagues (2013), whose methods we will now briefly summarize. They chose 500 common nouns, and all possible noun-noun combinations were generated. The occurrence of these combinations as two-word phrases was cross-referenced with a large corpus of human-generated text. Phrases with meaningful interchangeable word orders or that were taboo were removed. 'Nonsense' or low-meaningfulness phrases were generated by reversing the word order of meaningful phrases, resulting in 2,160 phrases.

Participants (N=150) rated subsets of the total phrase pool with the following instructions:

*Please read each phrase, then judge how meaningful it is as a single concept, using a scale from 0 to 4 as follows: If the phrase makes no sense, the appropriate rating is 0. If the phrase makes some sense, the appropriate rating is 2. If the phrase makes complete sense, the appropriate rating is a 4. Please consider the full range of the scale when making your ratings.*

*Examples: the goat sky, 0 (makes no sense), the fox mask, 2 (makes some sense), and the computer programmer, 4 (makes complete sense).*

For each phrase, the mean and standard deviation of participant responses were calculated. Here, 392 phrases with mean ratings between 1.5 and 2.5 were removed from the set due to being ambiguous to human raters, and resulted in 977 nonsense and 761 meaningful phrases used in the TWT presented here.

**The Two Word Test: Assessment of Combinatorial Semantic Understanding in LLMs**

We conducted a series of experiments comparing GPT-4, GPT-3.5-turbo, and Bard performance (each model as available in April 2023) to the human data. First, we gave the LLMs the same prompt used by Graves et al., followed by an enhanced version of the prompt. Then, we tested the LLMs on a binary version of the test (i.e., 'makes sense' / 'nonsense' judgment instead of numerical ratings).

1. **TWT: Numerical Meaningfulness Judgments**

For each LLM, we submitted the instructions and examples originally provided by Graves et al in subsets of randomized order. We repeated the instructions each time we submitted a subset (due to token restrictions) to ensure that errors were not due to memory limitations. Using Graves' original prompt resulted in the LLMs largely neglecting to use the 1 and 3 ratings, the two ratings not used as example cases in Graves' original prompt. To encourage the LLMs to use the full rating scale, we provided two additional examples in the instructions for scores of 1 and 3 (*the knife army*, 1 (makes very little sense) and *the soap bubble*, 3 (makes a lot of sense)). Compared to the human distribution, which reflects 'makes sense' and 'nonsense' phrases in the bimodal peaks, LLMs show a bias towards rating most phrases as a 2 or 3 (makes some sense, makes a lot of sense; Fig. 1).

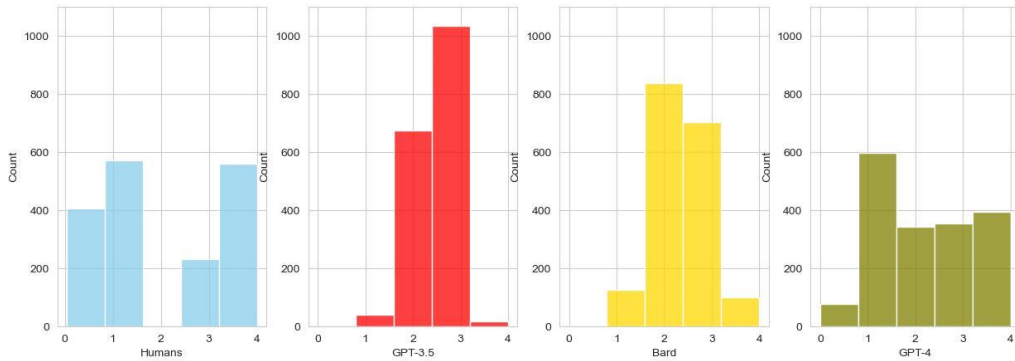

**Figure 1**  Frequency of continuous meaningfulness ratings for humans and LLMs. Human mean responses reflect a bimodal distribution of meaningful and nonsense phrases, while that is lacking in all three LLMs.

However, it is more informative to take LLM ratings of each individual phrase and test the probability that its rating came from the same distribution as the human responses to that phrase. We conducted a series of phrase-wise statistical tests to compare each LLM to human meaningfulness ratings.

First, we used the human phrase-wise means and standard deviations to generate a gaussian distribution of 10,000 simulated human responses to each phrase, respecting the lower and upper limits of the 0-to-4 scale and rounded to the nearest integer to match the LLM response scale (Fig. 2). Then, for each phrase, we conducted a Crawford & Howell t-test for case-control comparisons with the LLM as the case and the human distribution as the control. This modified t-test is designed for comparison of a single-case observation to a control group and returns the probability that the case comes from the same distribution as the group. We hereby define a 'TWT failure' as when the LLM meaningfulness rating has less than a 5% probability of coming from the human distribution (i.e., the LLM rating significantly different from that of humans, $p < .05$).

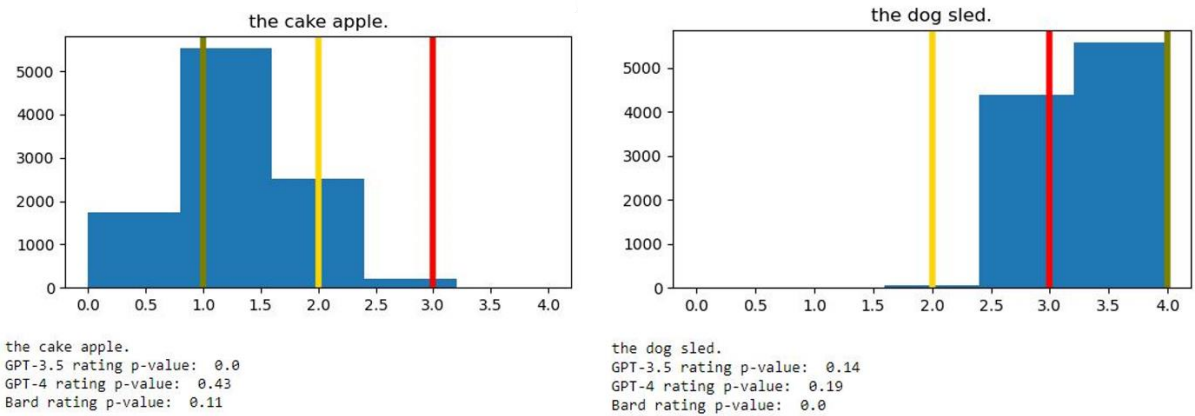

**Figure 2**: Simulated human rating distributions (blue) and LLM ratings (GPT-4 = olive, GPT-3.5 = red, Bard = yellow) for low- and high-meaningfulness phrases (*the cake apple*, *the dog sled*). For *the cake apple*, GPT-3.5 rated it as more meaningful than > 95% of humans would be expected to, while GPT-4 and Bard responded within normal limits. For *the dog sled*, Bard rated it as less meaningful than > 95% of humans would be expected to, while the other LLMs responded within normal limits.

We tested the LLMs with all 1,768 phrases, then limited it further to phrases most agreed-upon by human raters (determined by 95% confidence intervals around the human mean ratings). For 95% CI, there were 499 meaningful and 369 nonsense phrases. Table 1 provides TWT failure counts for the LLMs in the three subsets of phrases.

| LLM | % of failures: 1,768 phrases | % of failures, 868 most agreed upon phrases (95% CI) |
| --- | --- | --- |
| Bard | 42.7% | 57.9% |
| GPT-3.5 | 49.3% | 62.4% |
| GPT-4 | 23.4% | 23.2% |

**Table 1**     TWT failure percentages

To understand where LLM responses fall between human ratings and chance or random ratings, we generated two rating distributions. (1) 'Human': 1000 simulated participants whose phrase-wise responses were generated from the underlying probability distribution of responses to each phrase in the Graves et al. (2013) study. (2) 'Chance': 1000 permuted participants whose phrase-wise responses were selected based only on the frequency of 0-4 ratings from the original study. The 'Human' distribution approximates what would be expected from human raters if the study was run on a large number of human participants. The 'Chance' distribution is what would be generated by a system with no knowledge of word meaning. We then generated failure counts for the distributions and for each of the models.

**Experiment 1 Results:** Table 1 and Figure 3 show that Bard and GPT-3.5 failure counts are closer to chance than to the simulated human distribution. GPT-4 is significantly better than the other LLMs, but still fails far more than what would be expected from a human rater. Taken together, these results show that the three LLMs fail at the TWT, but that there are significant differences between their abilities.

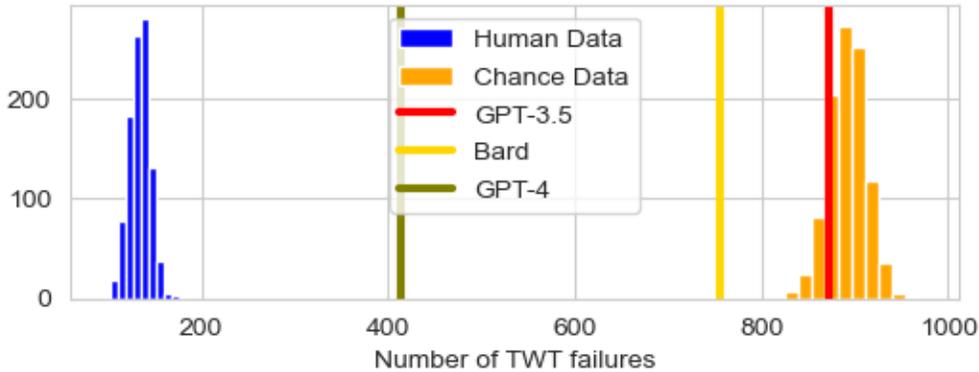

**Figure 3:** Number of LLM failures in TWT compared to simulated human (blue) and permuted-chance (orange) failure count distributions.

### 2. bTWT: Binary Meaningfulness Judgments

LLMs are often reported to make errors on numerical tasks. It is possible that the poor performance on the TWT was due to a difficulty in dealing with the numerical scale required for the task, rather than a lack of understanding of phrase meaning. In order to eliminate numerical ratings, we modified the TWT instructions to prompt binary responses:

*Please read each phrase, then judge how meaningful it is as a single concept. If the phrase makes no sense or makes very little sense, the appropriate response is 'nonsense'. If the phrase makes a lot of sense or complete sense, the appropriate rating is 'makes sense'.*

*Examples: 'the goat sky' is 'nonsense', 'the knife army' is 'nonsense', 'the soap bubble' is 'makes sense', 'the computer programmer' is 'makes sense'*

We then calculated the following to measure LLM performance: Chi-squared ($\chi^2$) test, signal detection theory (SDT) metrics, and receiver operating characteristic (ROC) curve.

**Experiment 2 Results:** Table 2 and Figure 4 show SDT results. SDT measures how well an actor (LLMs) can detect true signal (meaningful phrases) while correctly rejecting noise (nonsense phrases). It uses ratios of hits (true positives), correct rejections (true negatives), false alarms (false positives), and misses (false negatives). d' is a measure of overall ability to discriminate, with 0 being chance-level and >4 being close to perfect discrimination. β measures response tendency, or whether an actor prefers to say that a signal is present (liberal) or absent (conservative). Base 10 logarithm of β, reported here, is interpreted as < 0 being liberal and > 0 being conservative. We also display the ROC curve (Figure 5) and report area under the curve (AUC).

| LLM | All 1,768 phrases | 868 most agreed upon phrases (95% CI) |
|---|---|---|
| **Bard** | d' = 0.55<br>β = -0.11<br>AUC = 0.60 | d' = 0.74<br>β = -0.13<br>AUC = 0.63 |
| **GPT-3.5** | d' = 0.78<br>β = -0.36<br>AUC = 0.59 | d' = 1.23<br>β = -0.54<br>AUC = 0.65 |
| **GPT-4** | d' = 1.79<br>β = 0.17<br>AUC = 0.81 | d' = 2.58<br>β = 0.20<br>AUC = 0.90 |

**Table 2:** d', β, and AUC for LLM 'makes sense'/'nonsense' discrimination

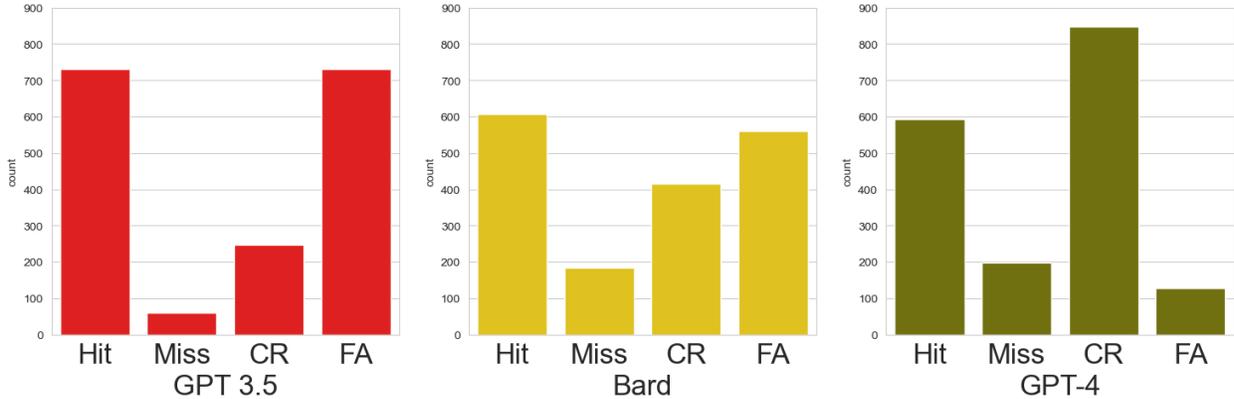

**Figure 4:** SDT metrics for all 1,768 phrases. Hit – true positive; Miss – false negative; CR – correct rejection (true negative); FA – false alarm (false positive).

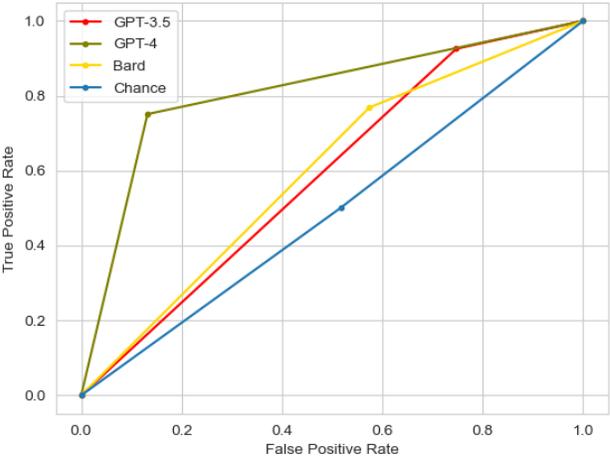

**Figure 5:** ROC curve for all 1,768 phrases

We also conducted the $\chi^2$ test. Briefly, $\chi^2$ test is used with categorical data and can test for statistical independence of observed frequencies to what is expected. Here, observed frequencies are the counts of LLM 'makes sense' and 'nonsense' responses and the expected response frequencies are those provided by the human data (e.g., 977 nonsense and 761 meaningful). Table 3 shows that the LLM frequency of responses are significantly different from the human response frequencies, and supports SDT and ROC results.

| LLM | $\chi^2$: all 1,768 phrases | $\chi^2$: 868 most agreed upon phrases (95% CI) |
|---|---|---|
| **Bard** | $\chi^2 = 84.3$ <br> $p < .001$ | $\chi^2 = 27.9$ <br> $p < .001$ |
| **GPT-3.5** | $\chi^2 = 538.1$ <br> $p < .001$ | $\chi^2 = 217.9$ <br> $p < .001$ |
| **GPT-4** | $\chi^2 = 10.6$ <br> $p = .001$ | $\chi^2 = 7.9$ <br> $p = .004$ |

**Table 3:** $\chi^2$ test results for observed (LLM) compared to expected (human) frequency of 'makes sense' and 'nonsense' responses. P < 0.05 indicates significantly different performance relative to humans.

**Results Summary:** Table 2 and Figures 4 and 5 show that Bard and GPT-3.5 display poor-to-modest discrimination. GPT-3.5 is overly liberal, tending to say that nonsense phrases make sense. Bard is more conservative and tends to say that sensible phrases are nonsense. GPT-4 is substantially better than the other models and displays moderate-to-high discrimination abilities. Almost the entirety of its improvement at the TWT over Bard and GPT-3.5 is by being able to correctly identify nonsense phrases. However, it is still significantly different from human performance.

**Conclusions and Future Work**

We presented a new benchmark for testing language understanding in LLMs. The task, essentially trivial for humans, requires rating meaningfulness of two-word phrases. Three current LLMs fail on this task. While GPT-4 performed better than GPT-3.5 and Bard, its performance still fell well short of humans. When test items were restricted to phrases that had the highest agreement among human raters, GPT-4 still provided statistically anomalous ratings on ~20% of phrases

A binary version of the test, bTWT, was created to test whether the poor performance of LLMs was the result of a failure to deal with the numerical scale required for TWT. The bTWT revealed that GPT-3.5 and Bard fail to distinguish meaningfulness of phrases binarily, achieving poor discrimination. GPT-3.5 was excessively liberal, tending to rate everything as 'making sense'. Bard was more conservative, often labelling sensible phrases as 'nonsense'. GPT-4, however, takes a significant step forward on the bTWT. While there is still room for improvement when tasked with judging all phrases, it displayed high discrimination abilities when probed on the phrases with low variability in human ratings.

Several investigations have begun to examine and reveal limitations of LLMs. For example, Dziri et al. (2023) tested LLMs on three compositional tasks (multi-digit multiplication, logic grid puzzles, and dynamic programming). They found that LLMs solve compositional tasks by reducing multi-step compositional reasoning into linearized subgraph matching. They suggest that in multi-step reasoning problems, LLM performance will rapidly decay with increasing complexity. Failures have been demonstrated in other problems as well, such as those involving logical and common-sense reasoning (Koralus et al.2023; Bian et al. 2023) was well as sequence tagging (Qin et al. 2023).

The TWT differs from these cases in that it does not directly require inference or reasoning. A limitation in breaking down a complex chain of reasoning into smaller problems should not affect performance on the TWT. Understanding these phrases requires understanding the constituent concepts, and then using world knowledge to determine whether the combination makes sense in some manner. A 'mountain stream' is a stream *located* on a mountain, but a 'stream mountain' is not a thing at all. An 'army knife' is not necessarily a knife located in the army but a type of knife useful in certain situations. TWT may exploit the fact that the text corpora that LLMs are trained on, no matter how large, almost entirely contain sensible text. However, this is the case for humans as well. Almost all text that people are exposed to is also sensible, but if the task requires, they are easily able to determine that certain word combinations don't make much sense. Current LLMs may lack the depth of real-world knowledge that is required for this task.

Many of the limitations of LLMs identified previously can be associated with a lack of 'executive control' that presents difficulties in complex symbolic or rule-based reasoning. Because of this, many have proposed combining deep neural networks with symbolic reasoning systems that can exert executive control when required (e.g., in three-digit multiplication). The weakness identified by TWT is qualitatively distinct, in that it is not directly related to the ability for executive control or systematic application of rules. It appears to be a limitation related to underlying semantic knowledge itself, rather than to reason using that knowledge.

These results also urge for caution in attribution AGI or similar abilities to LLMs, based on testing on tasks that are difficult for humans. The mounting understanding of the impressive abilities as well as limitations of LLMs will be essential in improving these models, and in identifying appropriate use cases.